\pdfoutput=1
\documentclass[letterpaper, 10 pt, conference]{ieeeconf}  

\IEEEoverridecommandlockouts                              

\usepackage{hyperref}					
\usepackage{microtype}				
\usepackage{cite}
\usepackage{amsmath,amsfonts} 
\usepackage[normalem]{ulem}
\usepackage{xpatch}
\usepackage{graphicx}
\usepackage{todonotes}
\usepackage{subfig}

\usepackage{multirow}
\usepackage{colortbl}

\title{\LARGE \bf
Robotic Irrigation Water Management: Estimating Soil Moisture Content by Feel and Appearance
}

\author{Marsela Polic, Marko Car, Jelena Tabak and Matko Orsag
\thanks{Authors are with Faculty of Electrical Engineering and Computing, University of Zagreb, Unska 3, 10000 Zagreb, Croatia
        {\tt\small  marsela.polic, marko.car, jelena.tabak, matko.orsag @fer.hr}}%
}

\begin{document}

\maketitle
\thispagestyle{empty}
\pagestyle{empty}

\begin{abstract}
In this paper we propose a robotic system for Irrigation Water Management (IWM) in a structured robotic greenhouse environment. A commercially available robotic manipulator is equipped with an RGB-D camera and a soil moisture sensor. The two are used to automate the procedure known as "feel and appearance method", which is a way of monitoring soil moisture to determine when to irrigate and how much water to apply. We develop a compliant force control framework that enables the robot to insert the soil moisture sensor in the sensitive plant root zone of the soil, without harming the plant. RGB-D camera is used to roughly estimate the soil surface, in order to plan the soil sampling approach. Used together with the developed adaptive force control algorithm, the camera enables the robot to sample the soil without knowing the exact soil stiffness a priori. Finally, we postulate a deep learning based approach to utilize the camera to visually assess the soil health and moisture content.
\end{abstract}

\section{Introduction}
The recent rise of robots in agriculture includes various application, ranging from picking \cite{xiong2020strawberry}, pruning \cite{billingsley2019agricultural},  pollination \cite{strader2019pollination} etc. However, taking care of the plant includes maintaining the health of the soil in which it grows. The \emph{feel and appearance method} is actually a well defined and proscribed procedure farmers use to schedule irrigation of their crops \cite{klocke1984g84}. It is a way of monitoring soil moisture to determine when to irrigate and how much water to apply. It is a common knowledge that plants need water to live and grow. However, applying too much water wastes this precious resource, and causes the loss of nutrients available for the plant. The feel and appearance of soil vary with texture and moisture content. Experienced farmers can estimate soil moisture conditions, to an accuracy of about 5 percent. Even though it is best to vary the number of sample sites and depths according to crop, field size, soil texture, and soil stratification, unfortunately to save time and effort the soil is typically sampled at three or more sites per field. For each sample the "feel and appearance method" involves various steps of tactile and visual inspection, comparing observations with photographs and/or charts to estimate percentage of the available water.

One of the goals within the SpECULARIA project \cite{specularia} is to automate procedures like this, reducing human labor input in small indoor farms by replacing it with a heterogeneous team of robots. This team of robots is used in structured greenhouse cultivation, where plants are grown in container units so that they can be transported around the greenhouse by an unmanned ground vehicle (UGV). The UGV transports the plants to the workstation, where a robotic manipulator treats the plants under controlled conditions. The structure of the controlled workspace around the robot gives it an upper hand when compared to mobile robots that manipulate plants in various conditions all over the farm.
\begin{figure}[t!]
	\centering \includegraphics[width=0.96\columnwidth]{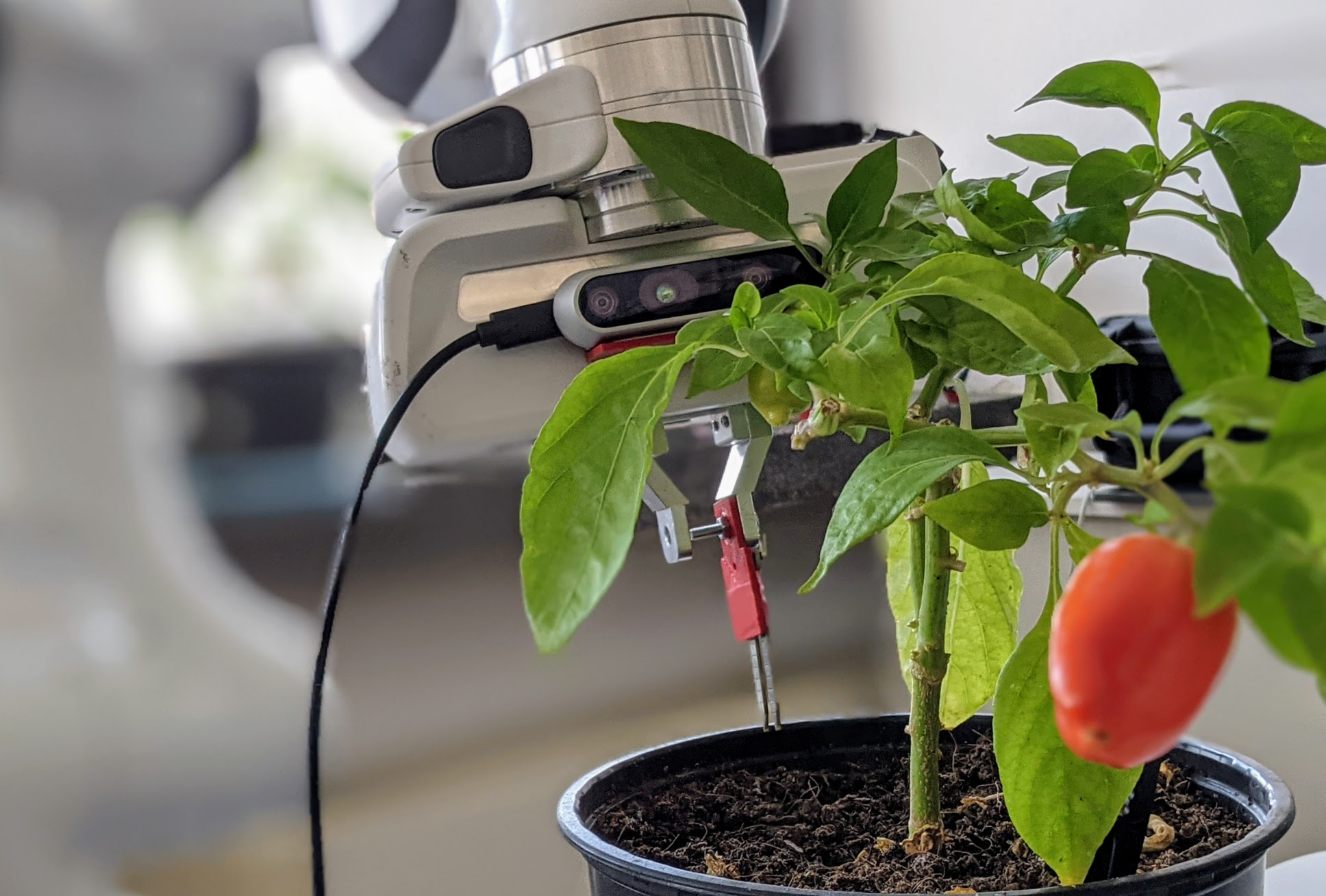} \caption{Collaborative robot Franka Panda performing soil moisture measurement procedure on a sweet pepper (Capsicum anuum) plant with a smart IoT moisture sensor.} \label{fig:franka_pepper_picking} \vspace{-0.55cm} 
\end{figure}
This paper focuses on the adaptive compliant control algorithm that enables the robot to sample the soil around the sensitive parts of the plant, and measure the water content close to the root of the plant. The method allows manipulation of objects of variable and unknown stiffness, ranging from manipulation of soft, wet ground, to handling collisions with rigid object such as roots or stones. The collaborative manipulator, which is the focus of this paper, is equipped with an RGB-D camera and a soil moisture sensor. The camera is used to estimate the position of the soil in the pot. Even though a rough estimate of the soil surface can be known a priori, the exact position varies during the entire vegetation process, as well as across the plant containers. Furthermore, the compliant control method additionally provides an estimate of the equivalent stiffness of the soil, thus providing another characteristic of the considered soil that can be used for describing the conditions. Combining different modalities will ultimately enable us to derive an AI based Expert system capable of soil moisture condition estimation utilized to plan optimal irrigation strategies for the greenhouse.

\section{Related work}

The work presented here relies on a widespread IoT solution for soil moisture measurement. The Soil Moisture Sensor utilizes a simple breakout for a straightforward method to measure water content in the soil \cite{saleh2016experimental}. The two exposed pads function as probes acting as a variable resistor. The amount of water in the soil is reflected in the electrical conductivity between the pads,  and is observed as a lower effective electrical resistance. The standard farming approach is to place the sensors and keep them in the ground for continuous measurements. Unfortunately, the exposed pads are quick to corrode, causing inconsistent measurements and harming the plants. Therefore, this sensor solution is recently being replaced with a more expensive versions of the soil moisture sensors. However, in the envisioned scenario we aim to utilize the existing resistance-based sensors, for two main reasons. First, such an end effector adapter is cost-effective, and the deployment with the robotic arm as opposed to fixed long-term measurements reduces the corrosion issue. The second motivation to use this type of sensor is in its physical resemblance to the U-fork, patented in France 1963, also known as the grelinette \cite{forkpatent}. This tool is intended for soil aeration and drainage performed by digging the dirt around the plant to gently loosen it.


Inspired by the pioneering work of Hogan \cite{hogan1984impedance}, impedance control of robotic manipulators has been extensively researched over the past decades, resulting in a development of a whole range of different impedance control strategies \cite{alshuka2018impedance}. Impedance controllers have been applied for a wide variety of tasks, including, but not limited to, robotic rehabilitation, industrial manipulation, micro-manipulation and agricultural grasping and picking \cite{song2019impedance}. It has been shown that both accurate force tracking and soft grasping can be achieved using impedance controller \cite{ting2014impedance, sano2013impedance}, which has been successfully tested in agricultural applications as well \cite{wang2015impedance, polic2021}. In \cite{wang2015impedance}, authors presented an impedance control strategy for compliant fruit and vegetable grasping, while an impedance-based compliant plant exploration framework has been presented in \cite{polic2021}. 
In this work, the classical impedance control is extended with an adaptive control law that enables the robot to sample the soil without a priori knowing its exact stiffness.

To enable the adaptation law to estimate the stiffness of the ground, however, prior to placing the sensor in the ground, it is necessary to estimate the exact position of the soil surface inside the pot. Common approaches to the plane detection problem include computing surface normals and using standard Random Sample Consensus (RANSAC) algorithm \cite{holz2012plane}, or its improved, noise-resistant version \cite{li2017plane}, on the point cloud obtained with an RGB-D camera. Alternative approaches are based on region growing methods \cite{anhvu2015plane} and Hugh transform \cite{hulik2014}, or a combination of both \cite{leng2015plane}. The approach proposed in this work uses the high performing RANSAC algorithm and addresses the problem of spurious planes by fitting the plane model only on the subset of the point cloud extracted with the custom ground plane detection method. This custom ground plane detection uses the existing information about the structured conditions in which the plants are grown.

\section{Ground plane detection}

As the plant is watered while it grows over a period of time, the surface of the ground surrounding the plant tends to gradually drop. This adds to the uncertainty of the motion control parameters. Even though a rough estimate of the surface of the soil can be assumed from the structured nature of the greenhouse and the plant containers, knowing the precise position of the surface allows for commanding the compliant control motion with a predetermined force, and using the adaptive control strategy to estimate the stiffness of the ground. The ground plane was detected using information from a consumer RGB-D camera, Intel RealSense D435, which was mounted on the manipulator in an \textit{eye-in-hand} configuration. The camera calibration procedure was conducted in an autonomous manner, as described in \cite{maric2020unsupervised}, yielding the transformation between the camera ($L_c$) and the flange ($L_f$) where the camera is mounted $\textbf{T}_f^c \in \mathbb{R}^{4 \times 4}$.


The obtained camera-flange transformation,  along with the known robot kinematics $\textbf{T}_0^f(\textbf{q})$ , is used to transform the point cloud measurments $\boldsymbol{\pi}_i$ from the local camera frame to the global reference frame $\textbf{p}_i = \textbf{T}_0^f(\textbf{q}) \cdot \textbf{T}_f^c \cdot \boldsymbol{\pi}_i$. Here we used $L_0$ to denote the base frame, and $\textbf{q} \in \boldsymbol{R}^n$ to denote the \textit{n} joints of the robot. For clarity we assume mapping between homogenoeus and Cartesian space is done implicitly when using vectors. Once the point cloud is transformed, NaN values and points which are out of the robot reach are filtered. As the robot is operating within the structured environment, the \textit{z} coordinate of the table top position $z_{TT}$, on which the pot is placed, as well as the height of the pot $z_{pot}$, is known a priori, enabling us to filter points $\textbf{p}_i \cdot \hat{\textbf{z}}_0<z_{TT}$ and $\textbf{p}_i \cdot \hat{\textbf{z}}_0 > z_{TT} + z_{pot} + \epsilon$. $\hat{\textbf{z}}_0$ denotes the \textit{z} axis of the base frame $L_0$. Finally, the remaining set of points $\mathbb{E}_0$ in the reachable environment of the robot is defined as:

\begin{equation}
\begin{gathered}
    \mathbb{E}_0 = \{ \textbf{p}  \in \{ \mathbb{PC}\} | x_{min} < x_{p} < x_{max}, \\
    y_{p} < y_{max}, \\
    z_{TT} < z_{p} < z_{TT} + z_{pot} + \epsilon\},
    \label{eq:environment}
\end{gathered}
\end{equation}
where the point $\textbf{p}$ from the point cloud $\mathbb{PC}$ belongs to the set of points $\mathbb{E}_0$ if it is within the robot reach and if its $z$ coordinate is bigger than the $z_{TT}$ and smaller than the sum of $z_{TT}$ and $z_{pot}$, increased by the small positive value $\epsilon$. The set of points $\mathbb{E}_0$ is sorted based on the ascending \textit{z} value.

\begin{figure}
	\centering
    \vspace{0.25cm}
	\includegraphics[width=0.96\columnwidth]{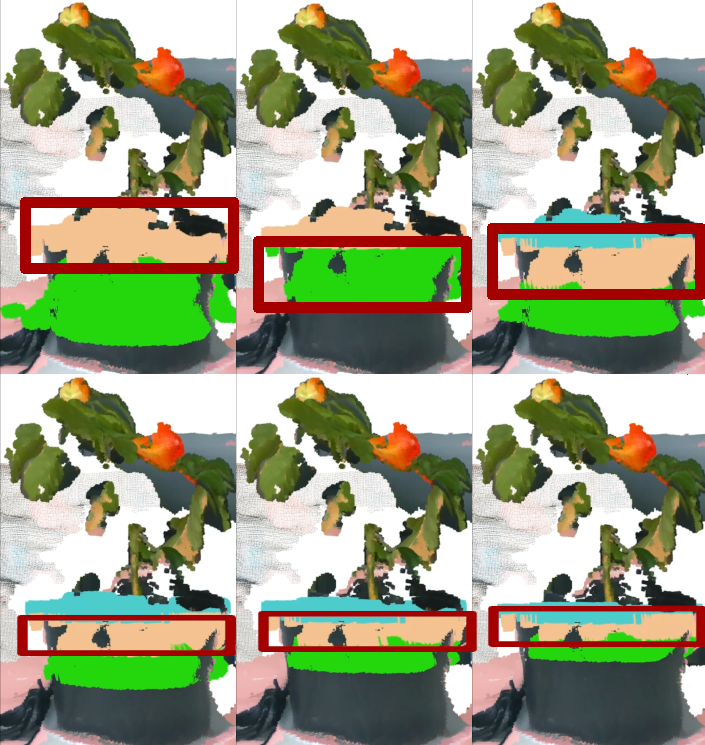}
	\caption{The ground plane is segmented by iteratively dividing the subset of points into bins, shown in different colors, and selecting the highest scoring bin (denoted with red rectangle).}
	\label{fig:bins}
\end{figure}

The points from the set $\mathbb{E}_0$ are split into bins based on their \textit{z} coordinates with the initial $\Delta z$ being set to $7cm$. Score for each bin is calculated according to the set of equations \ref{eq:score}.
\begin{equation}
\begin{gathered}
s1 = \dfrac{||B_{c-1}| - |B_c||}{|B_{c-1}| + |B_c| + 1} \cdot |B_c| \\
s2 = \dfrac{||B_c| - |B_{c+1}||}{|B_c| + |B_{c+1}| + 1} \cdot |B_c| \\
s = \dfrac{s1 + s2}{2},
\end{gathered}
\label{eq:score}
\end{equation}
where $|B_c|$ denotes the number of points in the current bin for which the score \textbf{s} is being calculated. The calculated score represents the number of points in the current bin, scaled with the relative difference between the number of points in the current bin and the number of points in the neighbouring bins. It is expected that the bin with the highest score contains the points which belong to the ground plane. After selecting the highest scoring bin, the set of points $\mathbb{E}_k$ is updated as in Eq. \ref{eq:env_star}:

\begin{equation}
    \mathbb{E}_k = \{ \textbf{p}  \in \{ \mathbb{E}_{k-1} \} | (z_{b_{min}} - \dfrac{\Delta z}{2}) < z_{p} < (z_{b_{max}} + \dfrac{\Delta z}{2})\},
    \label{eq:env_star}
\end{equation}
where $z_{b_{min}}$ and $z_{b_{max}}$ stand for the minimum and the maximum value of the $z$ coordinates of the points in the highest scoring bin, respectively. After each iteration, the value of $\Delta z$ is reduced by 25\%. The algorithm is terminated once the value of $\Delta z$ decreases to $1 cm$. The visualization of the bins throughout the single experiment is shown in Fig. \ref{fig:bins}.

Plane model is fitted on the remaining $\mathbb{E}_k$ set of points using open sourced Point Cloud Library \cite{pcl} implementation of the RANSAC algorithm \cite{ransac}. The center of the ground plane $\textbf{g}_{c}$ is defined as the median value across all three coordinates of the plane inliers and the point closest to the robot $\textbf{g}_{min}$ is obtained by replacing the $y$ value of the $\textbf{g}_{c}$ with the minimum value of the $y$ coordinates of the plane inliers.

Prior to conducting an experimental validation, the proposed method is validated in the simulation environment. Custom pepper plant models were generated for simulation validation, along with the plastic growth container and the soil surface inside it. The soil surface was modelled with variation along the $z$ axis for a more  realistic morphology. In the simulation environment, RGB and depth images of pepper plants are generated using 3D modeling software, Blender \cite{BlenderBook}. The generated images served as an input to the custom ROS package, \textit{blender\_rgbd\_ros} \cite{blenderPackage}, which converts them to the point cloud data and publishes both images and point clouds on separate ROS topics. Camera intrinsic parameters used for image generation correspond to the parameters of the camera used in the real experiments, Intel RealSense D435.

\begin{figure}
	\centering
	\includegraphics[width=0.96\columnwidth]{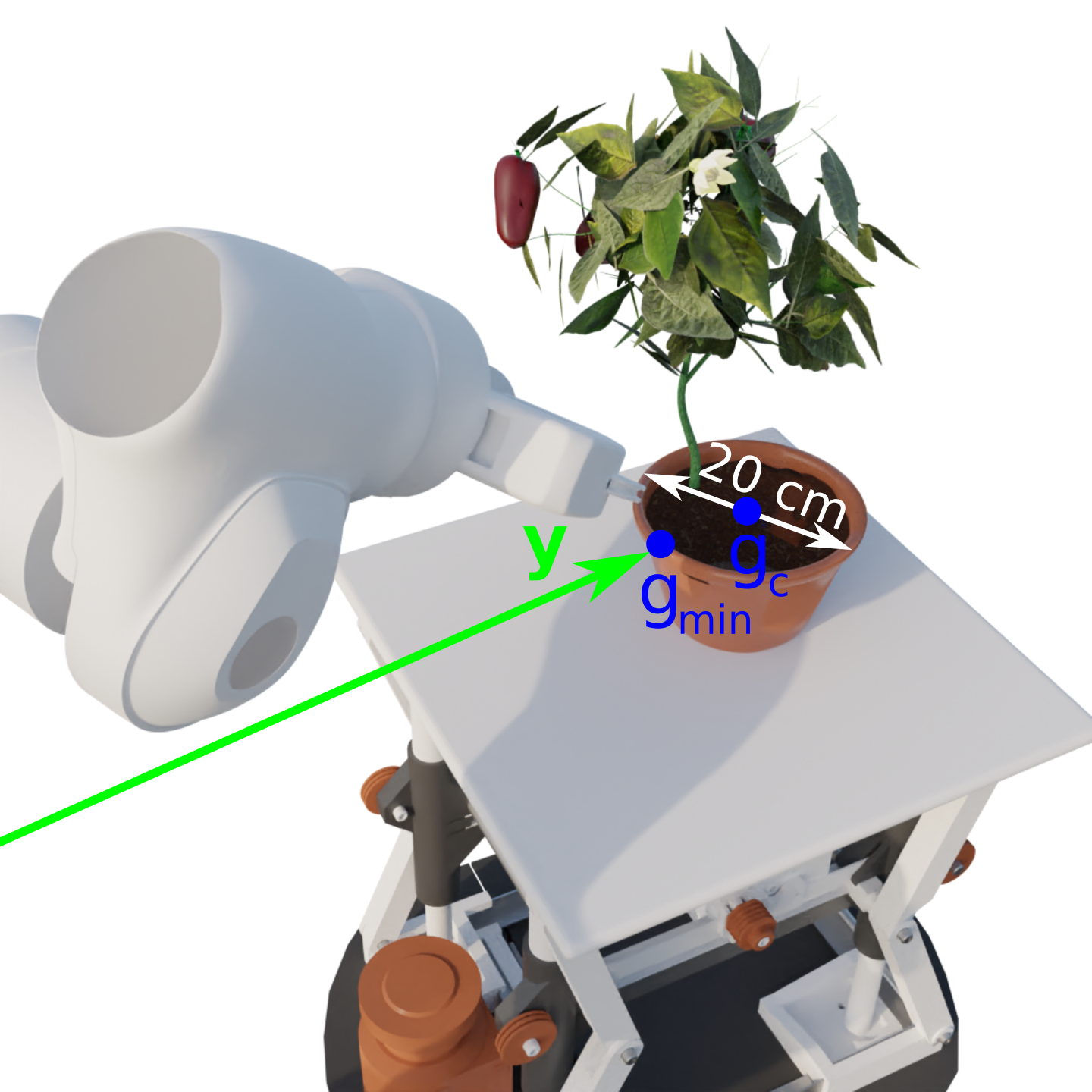}
	\caption{Blender setup for validation of the soil surface detection method. The simulated RGB-D camera records a realistic pepper plant model grown in a container. Robot approach vector $y$ is shown along with the ground truth point closest to the robot ($\textbf{g}_{min}$) and ground truth center ($\textbf{g}_{c}$).}
	\label{fig:blender_setup}
\end{figure}

\begin{figure}[h!]
	\centering
	\includegraphics[width=0.96\columnwidth]{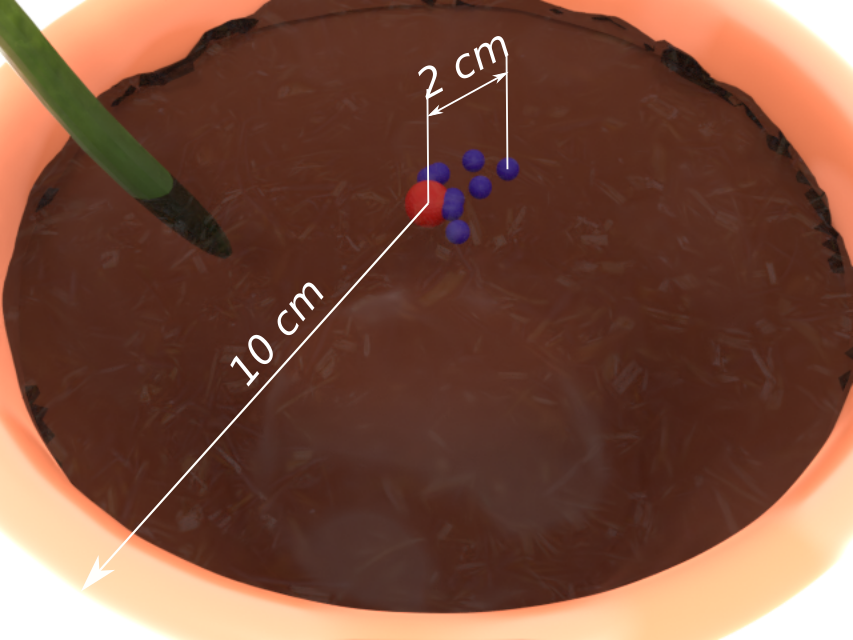}
	\caption{Visualization of the ground center estimates of the soil surface modeled in Blender. Red sphere denotes the ground truth center and the blue spheres denote the estimates for each of the 10 experiment repetitions. The radius of the modeled pot is $10cm$ and the maximum distance between the ground truth and the estimate is $2cm$. Texture of the soil surface is semi-transparent for the better visualization. }
	\label{fig:blender_estimates}
\end{figure}

Blender setup is shown in Fig. \ref{fig:blender_setup}. As in the real environment, $y$ axis of the reference coordinate system points towards the plant container. In the simulation environment, the exact position of $\textbf{g}_{c}$ and $\textbf{g}_{min}$ can easily be extracted, enabling the accurate comparison of the estimates and the ground truth. Both the mean value and the standard deviation of the soil surface estimates for a single plant recorded from 10 random positions, along with the ground truth value, are shown in Table \ref{tab:blender}. The error and the dissipation of the $x_{median}$, $y_{median}$ and $y_{min}$ are at a millimeter level, meaning that, even in the worst case scenario, estimated ground center lies within the close environment of the ground truth center, as can be seen in Fig. \ref{fig:blender_estimates}. The dissipation of the \textit{z} estimate, $z_{median}$, is even smaller, which was expected, considering that the plane model was fitted on the modeled surface plane points, which differed in the value of their \textit{z} coordinate for up to  $1cm$. The error and the dissipation of the \textit{z} estimate are sufficiently small for the successful implementation of the adaptive compliant control strategy. Knowing the central position of the soil with respect to the plant enables the robot to sample at the safe distance from plant's sensitive roots.

\begin{table}[h!]
    \vspace{0.25cm}
\centering
\caption{Ground truth and soil surface estimates in the simulation environment for the single plant recorded from 10 random positions. Diameter of the top of the pot is $20cm$.}
\label{tab:blender}
\begin{tabular}{|c|c|c|c|}
\hline
 & ground truth & mean & std\\
\hline
\hline
$x_{median}$ [mm] & 1.8 & 3.4 & 6.8 \\
\hline
$y_{median}$ [mm] & 0.8 & 4.0 & 5.0 \\
\hline
$y_{min}$ [mm] & -100 & -96.2 & 3.8 \\
\hline
$z_{median}$ [mm] & 110.2 & 111.2 & 0.1 \\
\hline
\end{tabular}
\end{table}

\section{Adaptive compliant control}

In this work, interaction of the end-effector with the environment relies on impedance control in the Cartesian space. The impedance filter models the robot-environment interaction system with an equivalent spring. The force tracking error between the desired contact force $\textbf{F}_r$ and the measured force $\textbf{F}$ as in eq. \ref{eq:forceError}, drives robot motion according to the desired mass-spring-damper system properties.
\begin{equation}
\label{eq:forceError}
\textbf{E} = \textbf{F}_r - \textbf{F}. 
\end{equation}
Here we consider force control along three spatial axes, so that all the vectors used are from $R^{3 \times 1}$. The user-defined target impedance behavior of the system determines the dynamic relationship between the robot position and the force tracking error so that it mimics a mass-spring-damper system as in eq. \ref{eq:targetImpedance},
\begin{equation}
\label{eq:targetImpedance}
\textbf{E} = \textbf{M} (\ddot{\textbf{X}}_c - \ddot{\textbf{X}}_r) + \textbf{B} (\dot{\textbf{X}}_c - \dot{\textbf{X}}_r) + \textbf{K} (\textbf{X}_c - \textbf{X}_r),
\end{equation}
where $\textbf{X}_c$ and $\textbf{X}_r$ are $R^{3 \times 1}$ commanded and reference position vectors of the end-effector, and $\textbf{M}$, $\textbf{B}$ and $\textbf{K}$ are the $R^{3 \times n}$ mass, damping and stiffness matrices of the target impedance, respectively. The reference position is the one provided by the user or by a higher level control. The commanded position is the actual input reference for the robot Cartesian position control. When in contact with the environment, the commanded robot position $\textbf{X}_c$ changes with the measured contact force through eq. \ref{eq:targetImpedance}, and the position tracking is in general not accurate. In other words, the impedance filter balances the position and force tracking errors. Switching to analysis along a single spatial axis without loss of generality, it can be shown  \cite{Seraji1993} that the desired contact forces can only be realized in case both precise environment position and the environment equivalent stiffness are known, by generating an adequate position reference $x_r$ using eq. \ref{eq:pos_ss}
\begin{equation}
\label{eq:pos_ss}
x_r = \frac{F_r}{k_e} + x_e,
\end{equation}
where $F_r$ is the desired contact force, and $k_e$ and $x_e$ are the environment stiffness and position, respectively. However, in the case presented in this work, the soil stiffness is unknown a-priori, and varies with the soil moisture. The controller used in this work is an extension of the classic position based impedance controller developed in \cite{contact2021markovic}, with online adaptation of the impedance filter inputs based on the estimated environment stiffness. The adaptation law for the position reference is based on the adaptive parameter $\kappa(t)$ that accounts for unknown elastic properties of environment under external force, 
\begin{equation} 
	x_r (t) =  \kappa (t) F_r + x_e,
\label{eqn:pos_kappa}
\end{equation}
where the position reference is a function of the estimate of the initial position of the environment $x_e$ and the force reference $F_r$. The exact adaptation law is given with eq. \ref{eq:adaptation} for one spatial dimension,
\begin{align}
k \dot{\kappa} (t) + b \ddot{\kappa} (t) + m \dddot{\kappa} (t) &= - \gamma_1 q (t) + \gamma_{1}^* \dot q (t), \nonumber \\
q(t) &= p_1 e(t) + p_2\dot{e}(t)
\label{eq:adaptation}
\end{align}
where $k$, $b$, and $m$ are the impedance filter parameters, $e(t)$ is the force tracking error, and $p_1$ and $p_2$ are the free parameters tuned based on the particular application. The derivation and convergence proof for this adaptation law can be found in \cite{markovic2021adaptive}. The adaptation law \ref{eqn:pos_kappa} used for ensuring force reference tracking implicitly yields a stiffness estimate of the manipulated object. In this case, the soil stiffness measure is used both for compliant manipulation in moisture measurement, and as a feature in the proposed soil moisture monitoring framework.
\section{Experimental Validation}
We conducted a series of experiments to verify how well the system estimates the position and the stiffness of the soil. In practice, the information gathered in such a way can help the system estimate the health of the soil and manage its water content.
\subsection{Ground plane detection precision}
\begin{figure}
	\centering 
	\includegraphics[width=0.96\columnwidth]{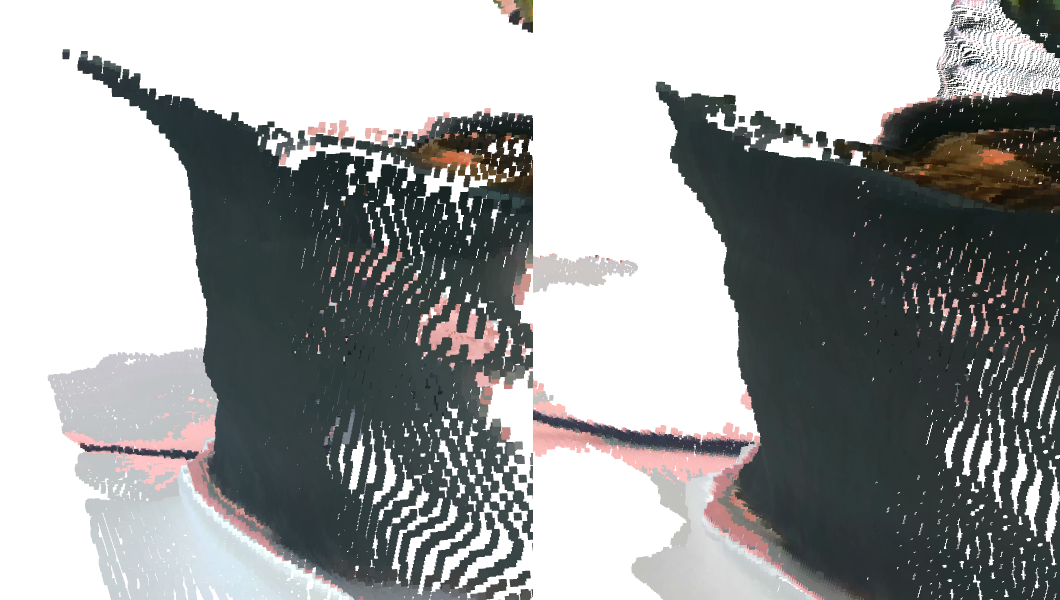} 
	\caption{Erroneous point cloud of pot recorded with Intel RealSense D435.} 
	\label{fig:camera_fail}
\end{figure}

In the same manner as in the simulation environment, a pepper plant was recorded from 10 random positions in the laboratory conditions. The center of the ground plane was estimated for each of the 10 frames as the median value of the fitted plane inliers across all three dimensions. The maximum difference and the dissipation of the estimates are shown in Table \ref{tab:real}. While the estimated value of \textit{z} coordinate of the ground center remained sufficiently small and robust to the change in the camera recording angle, the estimated \textit{x} and \textit{y} values significantly deteriorated compared to the simulation environment. This can be explained by the imperfections of the depth module of the RealSense camera. An example of an inaccurate point cloud of pot is shown in Fig \ref{fig:camera_fail}. This problem was addressed by extracting the closest point along the \textit{y} coordinate on the detected plane with respect to the robot base,  instead of the median value. The final \textit{y} coordinate of the reference position is defined as $3cm$ further along the $y$ coordinate, in order to reach the soil instead of the pot edge. As visible from the Table \ref{tab:real}, the dissipation is smaller when working with the $y_{min}$ instead of $y_{median}$. This was not the case in the simulation environment, as the camera model simulated in Blender did not incorporate the fluctuations in the depth data. However, considering that the diameter of the top of the pot is equal to $20cm$, dissipation in both  \textit{x} and \textit{y} values, though bigger than in the simulation environment, is still negligible. 

\begin{table}
\centering
    \vspace{0.25cm}
\caption{Dissipation of the estimates in the real environment for the single plant recorded from 10 random positions. Diameter of the top of the pot is $20cm$.}
\label{tab:real}
\begin{tabular}{|c|c|c|c|c|}
\hline
 & $x_{median}$ & $y_{median}$ & $y_{min}$ & $z_{median}$\\
\hline
\hline
$\Delta_{max}$ [mm] & 26.3 & 46.6 & 26.1 & 5.0 \\
\hline
std [mm] & 8.7 & 13.6 & 7.3 & 1.5 \\
\hline
\end{tabular}
\end{table}

\subsection{Control}



\begin{figure*}%
\centering
\subfloat[]{\includegraphics[width=.68\columnwidth]{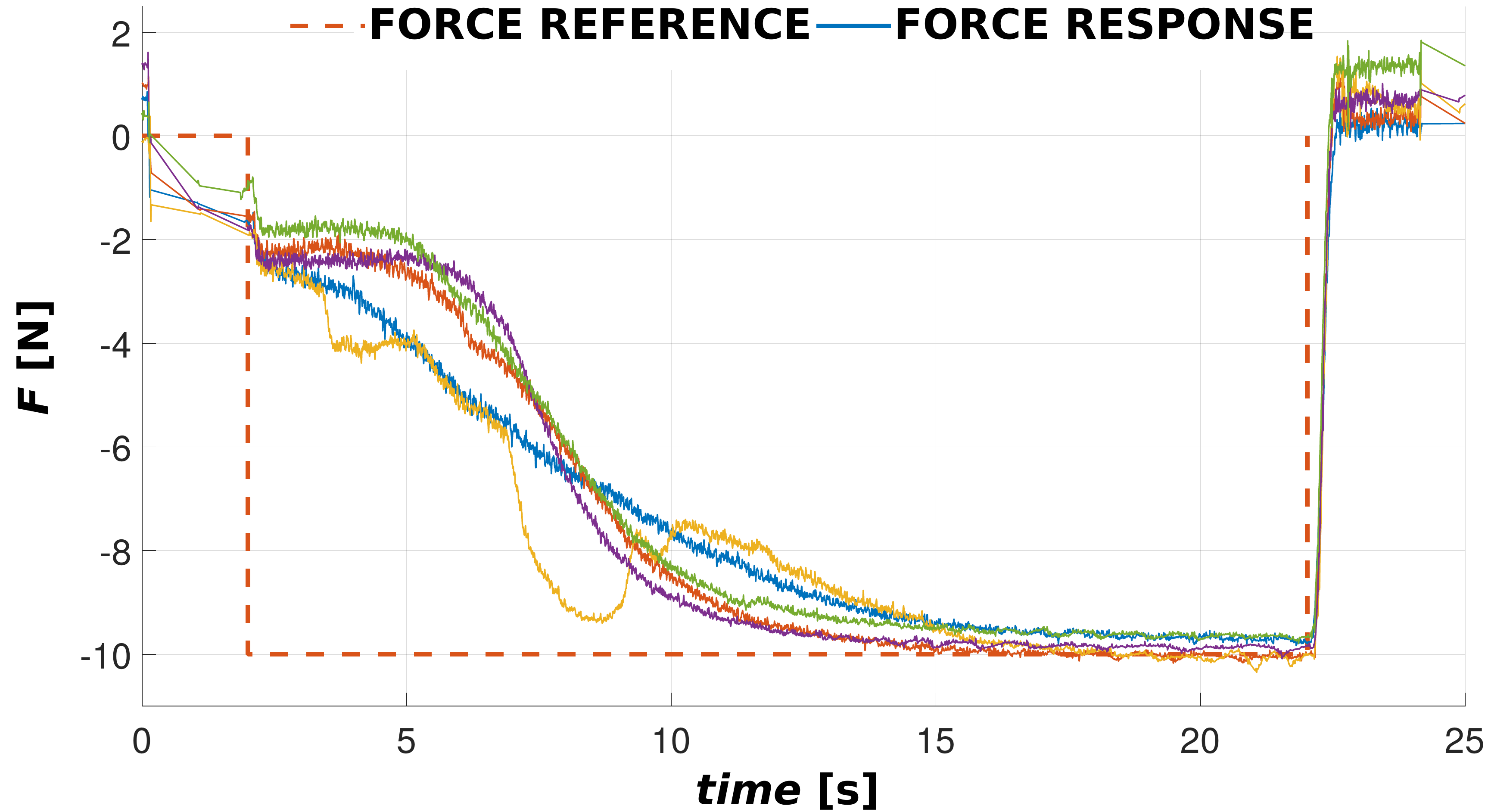} \label{fig:force_mokra5}} 
\subfloat[]{\includegraphics[width=.68\columnwidth]{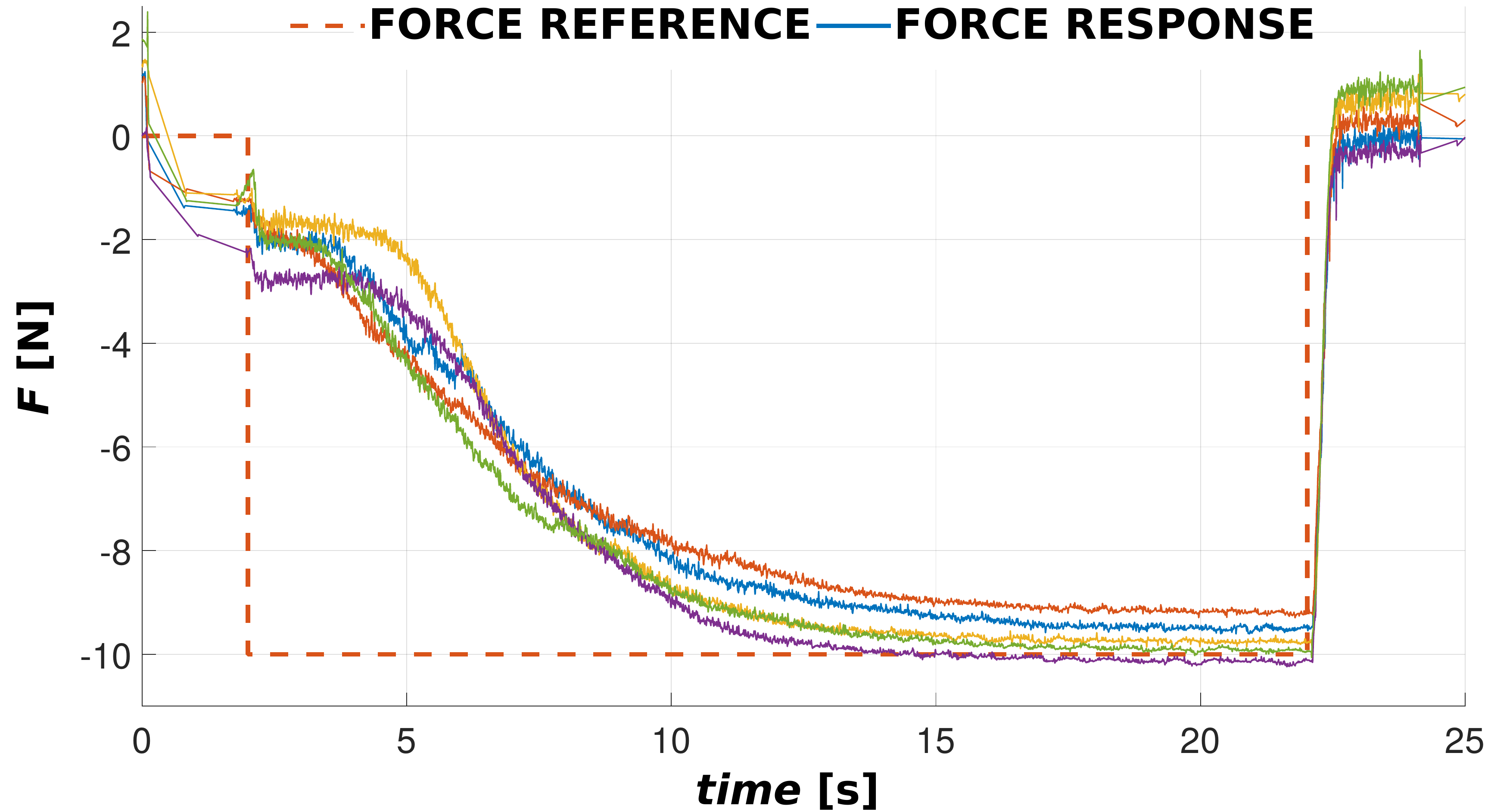} \label{fig:force_suha}}
\subfloat[]{\includegraphics[width=.68\columnwidth]{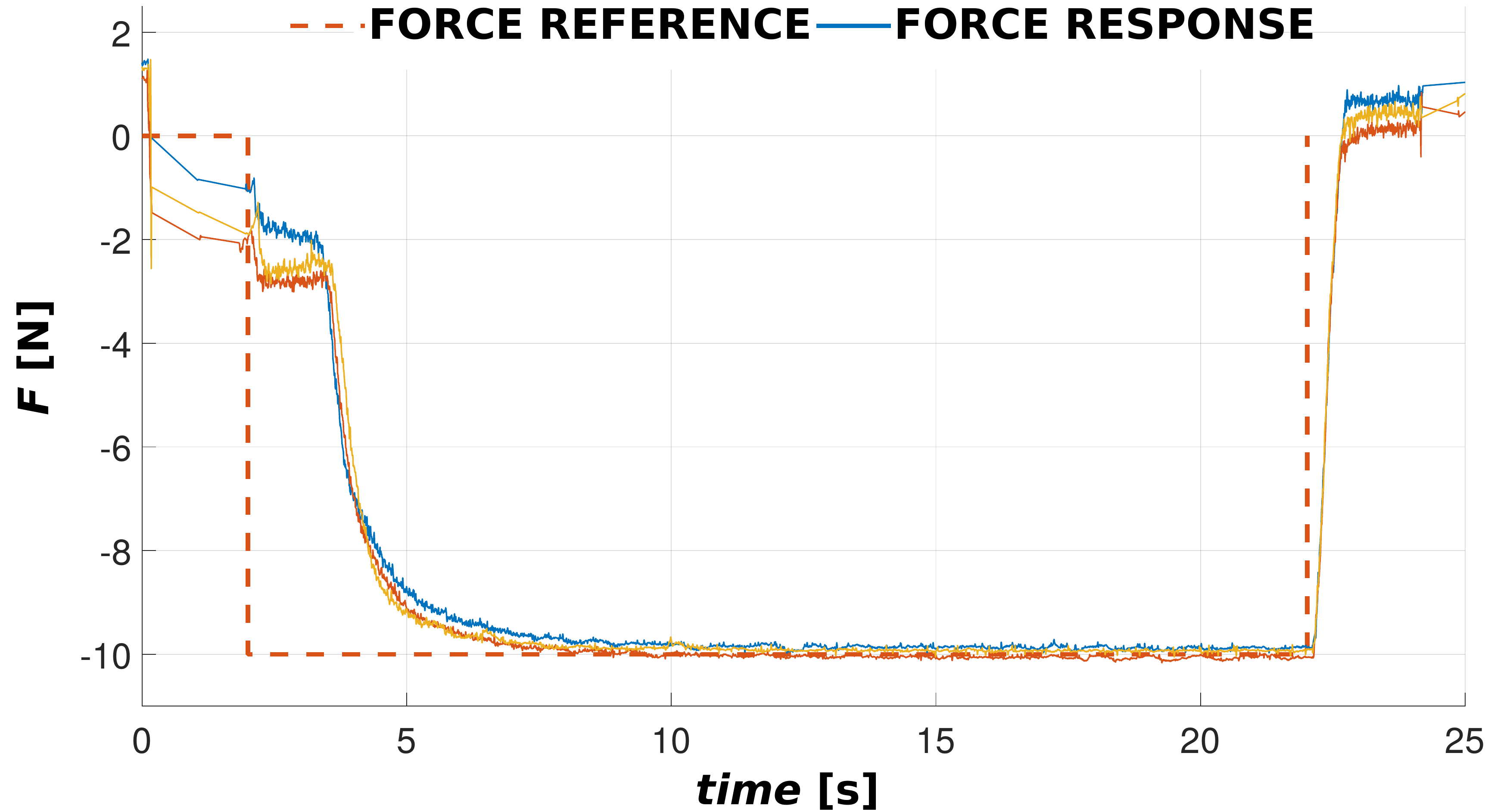} \label{fig:force_mokra3}}
\caption{Force tracking of the robot end effector in experiments on the softest object, moist soil \ref{fig:force_mokra5}, on the stiffer object, namely dry soil \ref{fig:force_suha}, and in case of collision with a rigid object \ref{fig:force_mokra3}. The robot motion is safe for the manipulated objects regardless of their stiffness.}
\end{figure*}

\begin{figure*}%
\centering
\subfloat[]{\includegraphics[width=.68\columnwidth]{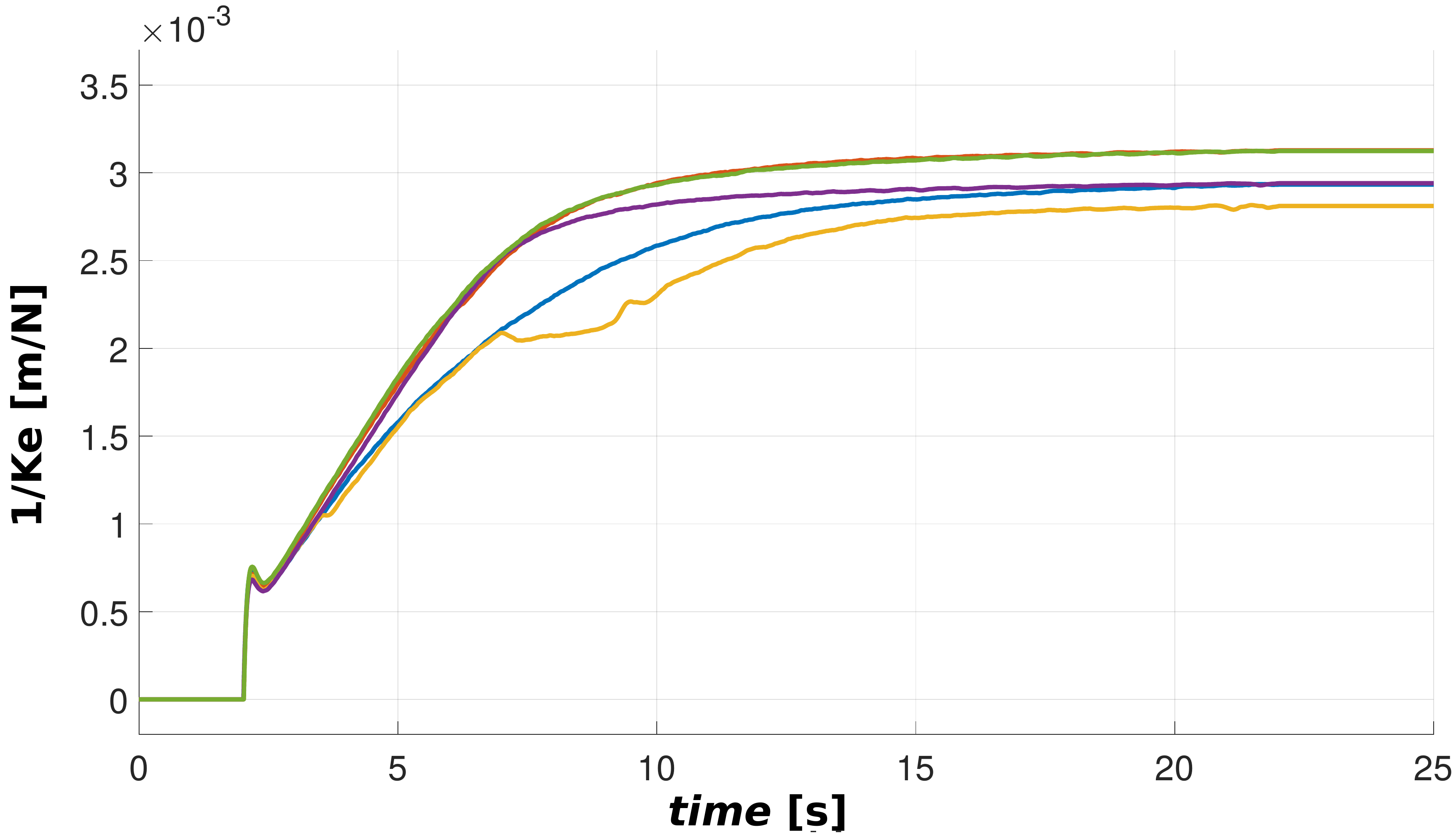} \label{fig:kp_mokra5}} 
\subfloat[]{\includegraphics[width=.68\columnwidth]{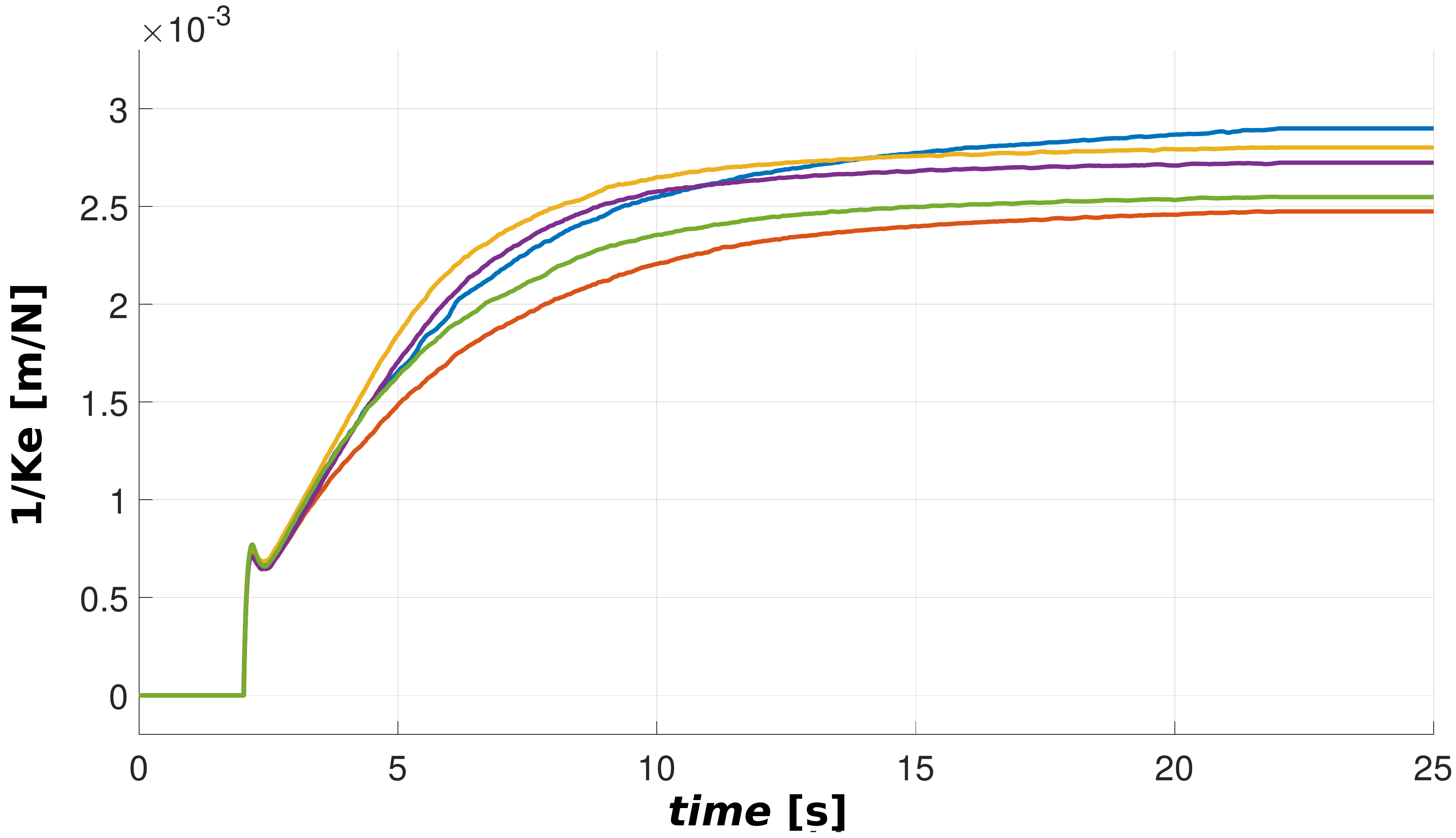} \label{fig:kp_suha}}
\subfloat[]{\includegraphics[width=.68\columnwidth]{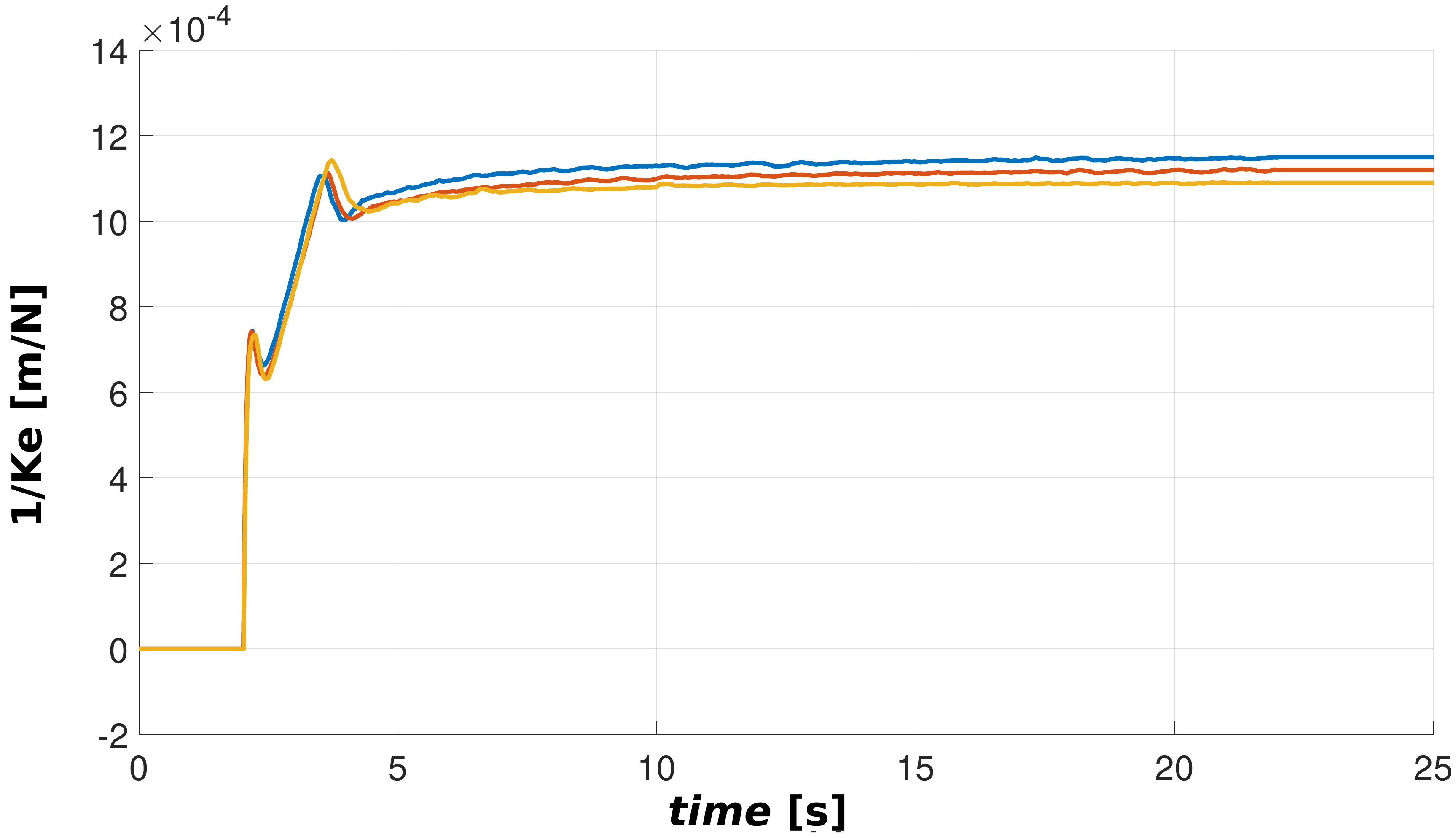} \label{fig:kp_mokra3}}
\caption{Adaptation dynamics during experiments with moist soil \ref{fig:kp_mokra5}, dry soil \ref{fig:kp_suha}, and in case of collision with a rigid object \ref{fig:kp_mokra3}. The estimation also compensates manipulator dynamics and surface detection pipeline. Adaptation tuned for fastest convergence on stiffest objects.}
\end{figure*}

Soil sampling experiments were conducted with a Franka Panda collaborative robot arm, based on the described soil surface detection using impedance control methods. The experiments were conducted in three scenarios analyzing the framework behavior with respect to different soil conditions. The moist soil conditions represent the softest scenario, i.e. the least stiff environment if considering the estimated stiffness values. The dry soil conditions represent a stiff scenario, while the extreme was tested through collision with a very rigid object, representing e.g. a stone in the soil that could potentially break the sensory equipment, or a part of the root system that should not be harmed. Five experimental repetitions were conducted for the first two scenarios, and three repetitions for the collision scenario. The results are represented in the graphs in figures \ref{fig:force_suha}-\ref{fig:kp_mokra3}. 

The force responses in the figures \ref{fig:force_suha}-\ref{fig:force_mokra3} are obtained with the same impedance filter, and the same parameters of the adaptation controller. The results show that the adaption in the framework is capable of reaching the desired contact force setpoint regardless of the stiffness of the manipulated object. When considering the  responses, it should be noted that instead of a precise external force/torque sensor, the measurement is provided by the Franka Panda dynamics estimation model. The imprecision in the model can be observed particularly in the variable baseline offset in the measurements at the beginning and at the end of each experiment. Here, the robot is not in contact with the environment, and there are no external forces acting on the end-effector. However, the estimated forces are non-zero due to model imprecision, and vary depending on the robot pose and velocity.




The adaptation framework implicitly models the stiffness of the manipulated object, as shown in \ref{fig:force_suha}-\ref{fig:force_mokra3}. Instead of the estimated stiffness, the responses show the dynamics of the inverse variable $1/K_e$, that could be considered compliance of the manipulated object (soil). For safety reasons, the initial assumption is that the manipulated object is infinitely stiff (zero compliance), and the adaptation of the estimated stiffness gradually reaches the actual value along with the desired contact forces. This is one of the reasons behind faster adaptation for a stiffer object, the other being the chosen adaptation parameters. These parameters would, in case of very compliant objects, probably have to be tuned for more aggressive (faster) adaptation. The compliance (stiffness) measure not only models the soil, but inherently takes into account the elasticity of the robot manipulator, as well as the imprecision of the soil surface detection. Regardless, the results show that throughout the repetitions, the estimation converges to the same region of values, proving the adaptation method is stable with respect to robot dynamics and detection imprecision. 

\section{Discussion}

This paper presents a soil moisture measurement method and its experimental validation applicable for robotic plant cultivation. The detection method in the initial step relies on an RGB-D visual setup for detection of the soil surface within the container. Instead of relying on color based segmentation, which is sensitive with respect to the lighting and other external conditions, or on other complex segmentation techniques such as CNNs, the method relies on an iterative 3D pointcloud segmentation and RANSAC based plane fitting. The method validation in simulation was confirmed in experiments with real plants, showing that the method is suitable for detection of soil plane inside the growth container. 

The actual measurement method relies on compliant control framework for a collaborative robot, that can easily be applied to industrial manipulator equipped with an external force torque sensor as well. The method enables compliant manipulator control during interaction with a deformable fragile objects, without prior information on the mechanical properties. Moreover, the method implicitly models them during motion control in attempt to reach the desired contact force setpoint. However, the obtained compliance measure not only models the soil but other system components as well, such as manipulator dynamics, and the error in soil surface estimation, which explains the variation in the estimations obtained over experiment repetitions. Even though not converging to precisely the same value over several experiment repetitions, the estimates of the compliance measures for three tested scenarios still distinguish between various manipulation conditions, and imply that such a measure could be used as a part of the feel and appearance method. Most importantly, the method is shown suitable and safe both for the robot, and for the underground plant parts, on a variety of soil conditions ranging from the soft moist soil to stiff soils with rigid debris in an autonomous manner without additional pre-tuning.

In the final setup, the visual part of the "feel and appearance method" will be achieved using the equipped RGB-D camera. Combining all three modalities (i.e. vision, measured resistance and stiffness) will enable us to derive an AI based Expert system capable of estimating soil moisture conditions to plan optimal irrigation strategies for the greenhouse.

\section*{ACKNOWLEDGMENT}
This work has been supported by Croatian Science Foundation under the project Specularia UIP-2017-05-4042 \cite{specularia2}.

\bibliographystyle{ieeetr}
\bibliography{root}

\end{document}